%% file: alad.tex
\documentclass[sigconf,anonymous=false]{acmart}
\AtBeginDocument{%
  }

\setcopyright{acmlicensed}
\copyrightyear{2018}
\acmYear{2018}
\acmDOI{XXXXXXX.XXXXXXX}
\acmConference[Conference acronym 'XX]{Make sure to enter the correct
  conference title from your rights confirmation email}{June 03--05,
  2018}{Woodstock, NY}
\acmISBN{978-1-4503-XXXX-X/2018/06}

\usepackage{multirow}
\usepackage{multirow}
\usepackage{tikz}
\usepackage{subcaption}
\usetikzlibrary{arrows.meta,positioning,fit,backgrounds}

\tikzset{
  blk/.style={draw,thick,rounded corners=1.5pt,minimum height=5.5mm,align=center,font=\small},
  froz/.style={blk,fill=black!8,draw=black!45,dashed},
  trn/.style={blk,fill=blue!12,draw=blue!65},
  emb/.style={draw,black!55,minimum width=3.6mm,minimum height=6mm,inner sep=0pt},
  ar/.style={-{Latex[length=1.5mm]},thick},
  lbl/.style={font=\scriptsize},
  tny/.style={font=\tiny},
}
\begin{document}

\title{Improving the Performance of Compressed Transformers with Post-Hoc Embedding Alignment}

\author{Rohit Raj Rai}
\email{rohitraj@iitg.ac.in}
\orcid{0009-0004-6002-1767}

\affiliation{%
  \institution{Indian Institute of Technology Guwahati
  \city{Guwahati}
  \state{Assam}
  \country{India}}
}
\author{Abhishek Dhaka}
\email{abhishek.dhaka7340@gmail.com}
\orcid{0009-0009-3978-9591}

\affiliation{%
  \institution{B.K. Birla Institute of Engineering \& Technology
  \city{Pilani}
  \state{Rajasthan}
  \country{India}}
}
\author{Amit Awekar}
\email{awekar@iitg.ac.in}
\orcid{0000-0003-1559-8247}

\affiliation{%
  \institution{Indian Institute of Technology Guwahati
  \city{Guwahati}
  \state{Assam}
  \country{India}}
}
\renewcommand{\shortauthors}{Rohit Raj Rai, Abhishek Dhaka, \& Amit Awekar}

\begin{abstract}
  Model compression is necessary for deployment of transformer models in resource-constrained environments. However, compression reduces the performance of transformer models on downstream tasks.  Existing solutions such as Knowledge Distillation (KD) and Parameter Efficient Fine-Tuning (PEFT) have their own limitations. KD trains the student under the teacher's supervision, but needs the teacher's training pipeline. PEFT methods such as LoRA adapt a model to a new task, but they do not recover the loss caused by compression. Our intuition is that a compressed model can be improved by aligning its token embeddings with those of a better performing reference model. We learn this alignment with a small feed-forward Alignment Adapter (AlAd) placed after the final layer of the compressed model. AlAd reads a window of neighbouring tokens, so it can use local context. AlAd is trained post-hoc, after the compressed model has been produced. The compressed model can either be kept frozen or jointly fine-tuned with AlAd. We study the effect of AlAd with BERT-base as the reference model and four compressed variants of the BERT family. We evaluate AlAd on three token-level tasks: POS tagging, NER and Extractive Question Answering. AlAd improves performance of compressed models on all three tasks. Joint fine-tuning gives a larger improvement in performance. The performance gain is the largest for the smallest models. AlAd provides a favorable trade-off between the efficiency and the performance of compressed models.
\end{abstract}

\ccsdesc[500]{Computing methodologies~Neural networks}
\ccsdesc[300]{Computing methodologies~Information extraction}
\ccsdesc[300]{Computing methodologies~Natural language generation}

\keywords{Deep Learning, Model Compression,  Embedding Alignment}


\maketitle

\section{Introduction}
\label{sec:intro}
Transformer models are compressed before deployment in resource-constrained environments such as edge devices~\cite{treviso2023efficient}.
Methods such as pruning, quantization, and knowledge distillation reduce the
size and the latency of a model~\cite{ganesh2021compressing}. However,
compression generally reduces the performance of transformer models on downstream tasks~\cite{ganesh2021compressing,turc2019}. The performance loss grows with the compression ratio. For instance, BERT-Tiny~\cite{turc2019}
retains about 4\% of the parameters of BERT-base~\cite{devlin-etal-2019-bert}. In our experiments we have observed that for BERT-Tiny, the Exact Match score on Extractive Question Answering (EQA) drops to just 15\%, against 74\% for BERT-base. An application must therefore
choose between efficiency and quality. This choice will affect more applications as edge deployment becomes more common.

Existing methods do not recover this loss for a model that has already been
compressed. Knowledge Distillation (KD) trains a small student model under the
supervision of a large teacher, matching its outputs or its intermediate
representations~\cite{hinton2015distilling,romero2015fitnets,jiao-etal-2020-tinybert}.
KD therefore produces a compressed model, but it cannot repair one. It also
assumes access to the teacher's training pipeline and a large training budget.
Parameter Efficient Fine-Tuning (PEFT) methods such as LoRA add a small number
of trainable parameters to a frozen model~\cite{hulora,houlsby2019parameter}.
They adapt a model to a new task. Recovering the loss caused by compression is
not their objective.

In practice, a compressed model is often already available, and it must be
improved as it is. Compressed models such as BERT-Mini, BERT-Tiny and
DistilBERT are released as pretrained artifacts and are downloaded
directly~\cite{turc2019,sanh2019distilbert}. A practitioner who deploys such a
model does not control how it was compressed. The teacher's training pipeline
and the compute budget it needs, are usually not available. Rebuilding the model
with a better compression method is therefore not an option. The only remaining
choice is to improve the compressed model after it has been produced. We call
this the post-hoc setting. In this setting, alignment is applied after
compression rather than during it. The compressed model may still be kept
frozen, or it may be fine-tuned jointly with the module that is added. This
paper asks how much performance can be recovered in the post-hoc setting, and
at what cost.

The post-hoc setting imposes three requirements on any solution. First, the solution must work for any compressed model, whatever method produced it. It cannot assume a particular pruning scheme, quantization scheme, or distillation
recipe. It must also work when the compressed model differs from the larger model in the number of layers, in the representation dimensionality, or in both. Second, the solution must work in augmentation with methods that are already applied
to the compressed model. A practitioner who has adopted LoRA or a distilled backbone should not have to abandon it. Third, the solution must improve the representation of every token, and not only the representation used for instance-level classification. Token-level tasks are those that depend on the
representation of every token in the input instance. The solution should improve the performance of the compressed model on token-level tasks. These three requirements follow from the fact that the compressed model is given rather than chosen.

Our intuition is that a compressed model can be improved by aligning its token embeddings with those of a larger model. The power of a neural model comes largely from the representations it learns for its input~\cite{bengio2013representation}. A transformer produces an embedding for every input token. These embeddings are the only interface between the model and the task-specific inference head. Compression reduces the depth or the width of the model, and therefore its representational capacity. Inputs that the larger model maps to well separated regions of its embedding space are mapped closer together by the compressed model. The inference head then has to operate on less separable embeddings. However, the compressed model might not discard all of this information. Part of it might be retained in a form that a learned transformation can recover. Such a transformation can push the embeddings of the compressed model closer to those of the larger model. 

\begin{figure}[t] 
\centering
\includegraphics[width=\columnwidth]{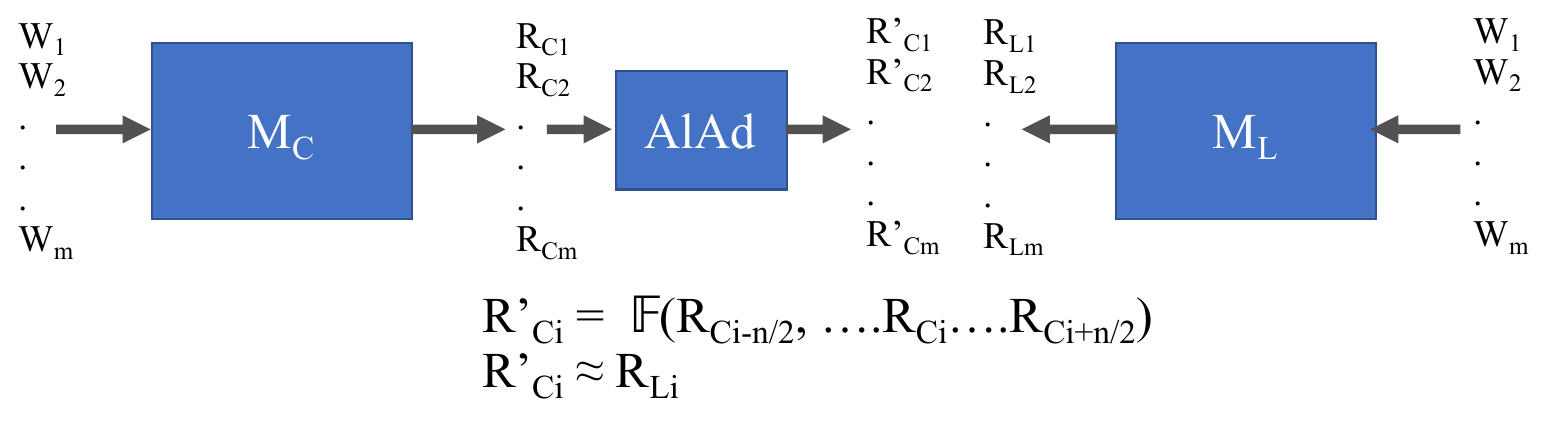}
\caption{Overview of the Alignment Adapter (AlAd) \newline
$M_C$ is a compressed model. $M_L$ is a reference model. Both receive input tokens $W_1$ to $W_M$. AlAd transforms embeddings of $M_C$ from $R_C$ to $R'_C$ to better align with $R_L$ (embeddings of  $M_L$). While computing the transform function $\mathbb{F}$, AlAd uses a window size of $n$ to utilize the context around each token.}
\label{fig:alad-overview-small}
\end{figure}

We realise this intuition as the Alignment Adapter (AlAd), a small feed-forward network placed after the final layer of the compressed model. AlAd takes the token embeddings produced by the compressed model and outputs embeddings in the representation space of the larger model. Figure~\ref{fig:alad-overview-small} shows an overview of the approach. AlAd is trained to minimise the distance between its output and the embedding that the larger model produces for the same token. The same transformation is applied at every token position. A transformation that reads a single position can only use the embedding at that position. AlAd therefore reads a window of $n$ neighbouring embeddings, which allows it to use local context while producing the embedding for a token. The compressed model may be kept frozen, or it may be fine-tuned jointly with AlAd. The first mode leaves the compressed model untouched. The second mode gives larger
improvements, and we report both results.

We study AlAd with BERT-base as the reference model and four compressed variants
of the BERT family. Table~\ref{tab:models} lists these models. They span 4\% to
60\% of the parameters of BERT-base. They are produced by three different compression paradigms. Two of them are not derived from BERT-base at all. We chose the BERT family of compressed models because their sub-100M parameter size makes them suitable for deployment in extremely resource constrained environment. We evaluate on three token-level tasks, namely POS tagging, Named Entity Recognition (NER), and EQA. When jointly fine-tuned, AlAd improves performance on all three tasks and on all
four compressed models. The improvement is largest for the smallest models. AlAd also works in augmentation with PEFT methods (LoRA) and compressed models created with KD (DistilBERT). AlAd increases the size of the compressed model and reduces its speedup. In return, a compressed model with AlAd becomes competitive with the next larger model in performance. It still has a smaller size and usually faster inference than the next larger model.

This paper makes two contributions. First, we present AlAd, a sliding-window
adapter that aligns the token embeddings of a compressed model with those of a
larger model. It satisfies the three requirements of the post-hoc setting,
because it uses only the output embeddings of the compressed model. Second, we
present a systematic study of post-hoc embedding alignment. We evaluate AlAd
across four compressed models, three compression paradigms, and three token-level
tasks. We explore two modes of training AlAd. We show that AlAd works in augmentation with LoRA and knowledge distillation. For complete reproducibility, all code, datasets, and trained models are available publicly\footnote{https://github.com/anonymouscode816/alignment-adapter}.

\begin{table}[t]
\centering
\caption{Models used in our experiments. Sizes are reported as a percentage of
the parameter count of BERT-base. $L$ is the number of encoder layers and $D$
is the representation dimensionality.}
\label{tab:models}
\small
\setlength{\tabcolsep}{4pt}
\begin{tabular}{l|cc|rr|l}
\hline
\textbf{Model} & \textbf{$L$} & \textbf{$D$} & \textbf{Params} & \textbf{Size}
& \textbf{Compression method} \\
\hline
BERT-base~\cite{devlin-etal-2019-bert}
 & 12 & 768 & 110M & 100\% & Reference model \\
\hline
DistilBERT~\cite{sanh2019distilbert}
 & 6 & 768 & 66M & 60\% & Distilled from BERT-base \\
ASC~\cite{10.1145/3703323.3703339}
 & 4 & 768 & 52M & 48\% & Layers pruned from BERT-base \\
BERT-Mini~\cite{turc2019}
 & 4 & 256 & 11M & 10\% & Pretrained at reduced scale \\
BERT-Tiny~\cite{turc2019}
 & 2 & 128 & 4.4M & 4\% & Pretrained at reduced scale \\
\hline
\end{tabular}
\end{table}

\input{figures/alad-figure}
\section{Alignment Adapter}
\label{sec:method}

We consider a compressed model $M_C$ and a reference model $M_L$. $M_C$ may be
derived from $M_L$, or it may be constructed independently. Let $D_C$ and $D_L$
denote the representation dimensionalities of the two models. We do not assume
that $D_C$ and $D_L$ are equal. We also do not assume that the two models have
the same number of layers. Given an input sequence of tokens $W_1, \dots, W_m$,
both models produce token-level embeddings. $M_C$ yields $R_{C1}, \dots, R_{Cm}$
and $M_L$ yields $R_{L1}, \dots, R_{Lm}$. On a downstream task $T$, $M_C$
typically performs worse than $M_L$. Our objective is to improve the performance
of $M_C$ on $T$. We introduce an Alignment Adapter (AlAd) that transforms each
$R_{Ci}$ into a new embedding $R'_{Ci}$ such that $R'_{Ci} \approx R_{Li}$. The
same transformation is applied at every token position. AlAd uses only $R_C$ and
$R_L$. It does not use the weights or the intermediate layers of either model.

AlAd is a feed-forward neural network with a single hidden layer.
Figure~\ref{fig:alad-overview} shows the details of our approach. AlAd reads a
window of $n$ neighbouring embeddings. With $n=1$, it uses only $R_{Ci}$ to
produce $R'_{Ci}$. With $n=3$, it uses $R_{Ci-1}$, $R_{Ci}$ and $R_{Ci+1}$. The
input layer therefore has $n \times D_C$ units and the output layer has $D_L$
units. The hidden layer uses a nonlinear activation. We apply zero-padding at
the beginning and the end of the sequence, so that the window is defined for
every token. AlAd computes the transformation $\mathbb{F}$:
$$R'_{Ci} = \mathbb{F}(R_{Ci-\lfloor n/2 \rfloor}, \dots, R_{Ci}, \dots,
R_{Ci+\lfloor n/2 \rfloor})$$

A transformation that reads a single position can only use the context that
$M_C$ has already encoded at that position. A transformer produces contextual
embeddings, so $R_{Ci}$ carries information about the neighbours of $W_i$. This
aggregation is performed by the attention layers of $M_C$. Compression reduces
the number of these layers or their width, so $M_C$ aggregates less context than
$M_L$. Information about the neighbours of $W_i$ may then be present at the
neighbouring positions and not at position $i$. A window of $n$ embeddings gives
AlAd direct access to this information. AlAd can use it while producing
$R'_{Ci}$. The cost is that the input layer grows by a factor of $n$. We report
the effect of the window size in Section~\ref{sec:favorabletradeoff}.

AlAd aligns the embeddings of the final layer. Feature-based distillation methods also align intermediate
layers~\cite{romero2015fitnets,jiao-etal-2020-tinybert,sun2019patient}. This requires a correspondence between the layers of the two models, and the correspondence is chosen when the compressed model is designed. In the post-hoc setting the compressed model is given. BERT-base has 12 layers, while the compressed models we use have 2, 4 or 6. Their representation dimensionalities also differ. There is therefore no correspondence to use. The final layer is the only interface that every compressed model shares with $M_L$. It is also the representation that the task-specific inference module consumes. We do not compare against alignment at intermediate layers, since the post-hoc setting therefore admits only final-layer alignment.

AlAd is trained in three phases. In the first two phases, AlAd is trained to
minimise the Mean Squared Error between $R'_{C}$ and $R_{L}$ over all tokens. We
use the Mean Squared Error because the target is a real-valued embedding vector
rather than a distribution over classes. The input embeddings are supplied by
$M_C$ and the target embeddings are supplied by $M_L$. Both models are frozen in
these two phases. $M_L$ is frozen because it is only a source of target
embeddings. $M_C$ is frozen because AlAd learns a mapping from the output space
of $M_C$ to that of $M_L$. If $M_C$ were updated during these phases, its output
space would change, and the mapping learned so far would no longer apply. The
first phase is task-independent pretraining. AlAd learns the mapping on a large
general-purpose corpus, without reference to any task. The second phase is
task-specific alignment. AlAd continues to learn the same mapping on the
training set of $T$, and no task-specific inference is performed. The two phases
differ in the distribution of the input embeddings. A general-purpose corpus and
a task dataset differ in vocabulary, in sentence structure, and in domain, so
$M_C$ produces different embeddings for the two. The first phase learns a
mapping that holds broadly, and the second phase refines it where the task
actually operates.

The third phase is task-specific fine-tuning. AlAd is attached to $M_C$ together
with a task-specific inference module, and the combined model is trained with a
task loss. The inference module consumes $R'_{C}$ rather than $R_{C}$, so it
operates in the representation space of $M_L$. For POS tagging and for NER, the
inference module is a token classification layer that assigns a label to every
position. For EQA, it is a span prediction layer that scores every position as
the start and as the end of the answer. 


During task-specific fine-tuning, the compressed model may be kept frozen or
fine-tuned jointly with AlAd. In the frozen mode, only AlAd and the inference
module are trained. The parameters of $M_C$ are the same before and after. In
the joint mode, $M_C$ is trained together with AlAd and the inference module.
Both modes are post-hoc, since neither of them repeats the compression of $M_C$.
The two modes differ in what a practitioner must be able to do. The frozen mode
requires only that $M_C$ can be run in the forward direction. The joint mode
requires that $M_C$ can be trained. We report results for both modes in
Section~\ref{sec:results}.

AlAd can also be attached to a compressed model that is adapted with a PEFT
method. We use LoRA for this purpose. Low-rank matrices are injected into the
attention layers of $M_C$, and AlAd is attached after its final layer. The two
modules are trained at different times. The first two phases train AlAd against
a frozen $M_C$, as before, and the low-rank matrices are not used. The third
phase trains AlAd, the low-rank matrices, and the inference module together,
while the remaining parameters of $M_C$ stay frozen. The low-rank matrices
therefore change the output of $M_C$ after the alignment has been learned. This
combination is a way of adapting a frozen compressed model without fine-tuning
all of its parameters.

We also evaluate a linear adapter. It maps $R_{Ci}$ to $R'_{Ci}$ with a single
matrix. It has no hidden layer, no nonlinear activation, and no window. The
linear adapter is trained in the same three phases as AlAd, and it is fine-tuned
in the same way. The two differ only in the transformation they apply. A
comparison between them therefore indicates whether the hidden layer and the
window are necessary. We treat this as an empirical question and report the
answer in Section~\ref{sec:results}. One possible architecture for AlAd is to use attention mechanism. Attention over the whole sequence would let the adapter use context
beyond a fixed window. However, its cost grows quadratically with the sequence length.
In contrast, the cost of our sliding window-based design is independent of the sequence length. It grows linearly with window size. AlAd is retained at inference, so we prefer the design whose cost does not grow with the sequence length.

\subsection{Experimental Setup}
\label{sec:setup}

We use BERT-base as the reference model $M_L$. Table~\ref{tab:models} lists the
four compressed models. They span 4\% to 60\% of the parameters of BERT-b, and
they are produced by three different compression paradigms. DistilBERT is
distilled from BERT-b. ASC is obtained by pruning layers from BERT-b. BERT-M and
BERT-T are pretrained independently at a reduced scale, and they share no
weights with BERT-b. We chose these four models so that the evaluation covers
distillation, pruning, and reduced-scale pretraining, and so that it covers a
wide range of compression ratios. For BERT-M and BERT-T, AlAd maps representations between
two representation spaces that were trained separately. It is a harder problem as compared to training AlAd for DistilBERT and ASC.

We evaluate on three token-level tasks. A token-level task requires a prediction
at every position, or a prediction that depends on all positions, so it exercises
the entire output of the adapter. We chose three tasks that differ in what they
demand of the representation. POS tagging assigns one label from a small tag set
to every token, and the label depends mostly on the token and its immediate
neighbours. NER assigns entity types, and it depends on longer spans of context.
We use MultiCoNER v2, which is fine-grained and has 33 entity
types~\cite{fetahu-etal-2023-multiconer}. Many of these types are rare, so the
dataset also measures how well a model handles classes with few examples. EQA
requires the start and the end of an answer span to be identified together, so a
correct prediction depends on agreement across positions. We use SQuAD
2.0~\cite{rajpurkar2018know}, which also contains questions with no answer in
the passage. 


For POS tagging we use the German Universal Dependencies treebank
dataset~\cite{mcdonald-etal-2013-universal}, with 14,118 training, 799
validation, and 977 test sentences. All the models we use are pretrained on English, so this dataset places them in a cross-lingual setting. Compression is harder in this setting, because the compressed models retain less of the transfer ability of BERT-base. The first phase of alignment also uses an English corpus, while this task is in German. The alignment learned in that phase therefore cannot be a transfer of task knowledge from BERT-base. It can only be a mapping between the two representation spaces. MultiCoNER v2 has 16,778 training, 871 validation, and 249,980 test sentences. SQuAD 2.0 has 130,319 training and 11,873 validation
questions. For POS tagging we report accuracy. For
NER we report macro F1 score. Macro F1 is the official metric of MultiCoNER v2. For EQA we report exact match score. 

In our implementation, AlAd has a hidden layer of 1280 units with a GeLU activation. We evaluate window sizes $n \in \{1, 3, 5\}$. The linear adapter is a single $D_C \times D_L$
matrix with no hidden layer and no activation. For the PEFT experiments we use
LoRA adapters with rank $\textit{r} = 8$ and scaling factor $\alpha = 16$, introduced into the query and value projection matrices of each self-attention layer of $M_C$. A LoRA dropout of 0.05 was used, while no additional bias parameters were trained.

In the first phase, we perform general pretraining using the English Wikipedia corpus from March 2024 \footnote{https://dumps.wikimedia.org/}. We create three segments from the corpus each containing random 20 Million sentences. We sample 0.5\% of the training dataset into validation dataset and 0.5\% of the remaining training partition is used as test dataset. On each segment we perform pretraining for a total of 4 hours that is equivalent to 12 hours of pretraining. Validation is performed after every 30 minutes of pretraining. We use a learning rate of 5e-4 and AdamW as the optimizer. The objective of this pretraining is to align token-level embeddings of the compressed model, similar to that of BERT-base, using the Mean Squared Error (MSE) loss function. 

In the second phase, we perform continual pretraining of AlAd using the train partition of the downstream task with a slightly lower learning rate compared to what was used in the previous phase (1e-4). This pretraining is performed for a total of ten epochs with the same training objective as in the previous phase. During this phase, the adapter learns the task-specific token representation. 

Finally, in the third phase, the pretrained adapter is attached to the compressed model, and together they are fine-tuned to generate task-specific labels. We have two cases: in the first case compressed model remains frozen and only the adapter weights get modified. We choose a learning rate of 1e-4 in this case. In the second case the compressed model is also fine tuned along with the adapter. We choose a lower learning rate of 5e-5 because more parameters are being
updated, and they include pretrained backbone weights. A high learning rate can destroy this knowledge through catastrophic
forgetting~\cite{mccloskey1989catastrophic, kirkpatrick2017overcoming}. Sun et al. show that this occurs when BERT is fine-tuned, and that a lower learning rate mitigates it~\cite{sun2019finetune}. All experiments are run on a single NVIDIA A100 GPU with 80 GB of memory. In our experiments, the batch size was 32 and the maximum sequence length was 512 tokens.

\section{Results}
\label{sec:results}
BERT-base achieves an accuracy of 93\% on POS tagging, a macro F1 of 51\% on NER, and an exact
match of 74\% on EQA. In the rest of this section, we report all results as a percentage of the performance of our reference model BERT-base. A performance value of 100 in Table~\ref{tab:main} therefore means that the compressed model has retained the full performance of BERT-base on that task. A performance value greater than 100 in Table~\ref{tab:main} indicates that the performance of a particular configuration exceeds the performance of BERT-base on that particular task. In this paper, we report one metric per task. We also computed a second metric for each task, namely overall F1 for POS tagging, overall F1 for NER, and the token-level F1 of SQuAD for EQA~\cite{rajpurkar2018know}. The trends are the same on these additional metrics. The complete results are available in our repository. 


\begin{table}[t]
\centering
\caption{AlAd in joint fine-tuning mode improves performance of each compressed model for all three tasks. All performance values are a percentage of the performance of BERT-b
on the corresponding metric. Size is the size as a
percentage of BERT-base, and Speedup is the inference speedup over BERT-base.}
\label{tab:main}
\small
\setlength{\tabcolsep}{4pt}
\renewcommand{\arraystretch}{1.15}
\begin{tabular}{ll|ccc|cc}
\hline
\textbf{Model} & \textbf{Variant} & \textbf{POS} & \textbf{NER} & \textbf{EQA}
& \textbf{Size} & \textbf{Speedup} \\
\hline
\multirow{3}{*}{DistilBERT}
 & fine-tuned      & 100 & 88  & 89 & 60.00 & \textbf{1.60} \\
 & + Linear        & 100 & 102 & 91 & 61.48 & 1.58 \\
 & + AlAd          & \textbf{101} & \textbf{106} & \textbf{93} & 65.89 & 1.54 \\
\hline
\multirow{3}{*}{ASC}
 & fine-tuned      & 96 & 53 & 68 & 48.00 & \textbf{2.12} \\
 & + Linear        & 99 & 75 & 70 & 49.01 & 2.10 \\
 & + AlAd          & \textbf{100} & \textbf{80} & \textbf{74} & 53.89 & 2.01 \\
\hline
\multirow{3}{*}{BERT-M}
 & fine-tuned      & 92 & 27 & 61 & 10.20 & \textbf{2.50} \\
 & + Linear        & 96 & 51 & 59 & 10.45 & 2.46 \\
 & + AlAd          & \textbf{98} & \textbf{80} & \textbf{65} & 12.68 & 2.41 \\
\hline
\multirow{3}{*}{BERT-T}
 & fine-tuned      & 79 & 8  & 20 & 4.01 & \textbf{2.80} \\
 & + Linear        & 86 & 24 & 20 & 4.13 & 2.60 \\
 & + AlAd          & \textbf{91} & \textbf{59} & \textbf{35} & 5.70 & 2.27 \\
\hline
\end{tabular}
\end{table}

\subsection{Performance Improvement with AlAd}
AlAd improves each compressed model on all three tasks that we experimented with. We can observe that for each pair of model and task, AlAd beats the compressed fine-tuned model as well as a simple linear adapter. The improvement is larger for the more compressed models. Table~\ref{tab:main} reports the results with joint fine-tuning and window size of 5. For DistilBERT the POS accuracy rises by one point, and for
BERT-T it rises by twelve points. The largest improvements are on NER. The macro F1 of BERT-T rises from 8 to 59, and that of BERT-M rises from 27 to 80. Macro F1 weights every entity type equally, so this indicates that AlAd recovers
the performance on the entity types that the compressed model handles worst. AlAd also outperforms a linear adapter on every model and every task. The two adapters are trained in the same way and differ only in the sliding window and the non-linearity. The margin between them grows with the compression ratio. For DistilBERT it is at most four points, and for BERT-T it is up to 35 points. DistilBERT with AlAd is the only configuration that exceeds the performance of BERT-base, and it does so on POS tagging and on NER.

The benefit of AlAd grows with the compression ratio. DistilBERT retains 60\% of the parameters of BERT-b, and AlAd improves its POS accuracy by one point. BERT-T retains 4\%, and AlAd improves its POS accuracy by twelve points. The same ordering holds on NER and on EQA. Between the linear adapter and AlAd, the margin is at most four points for DistilBERT and up to 35 points for BERT-T. As compared to the linear adapter, AlAd is therefore more useful for models with higher compression.

\subsection{AlAd with Knowledge Distillation}

AlAd does not compete with knowledge distillation. It is applied to the model that distillation produces, and the two can be used together. DistilBERT is a model produced by knowledge distillation, and AlAd improves it on all three tasks (Table~\ref{tab:main}). DistilBERT with AlAd is the only configuration in our experiments that exceeds BERT-base. It reaches 101 on POS tagging and 106 on NER, so the distilled model with an adapter performs better than the model it was distilled from. The improvement over DistilBERT alone is smaller than for the other compressed models, because DistilBERT already retains most of the performance of BERT-b. A practitioner who has already distilled a model does not have to choose between the two methods. AlAd recovers performance that distillation alone did not.

\begin{table}[t]
\centering
\caption{AlAd can be used in augmentation with LoRA to improve model performance. All performance values are a percentage of the performance of BERT-base on the corresponding metric. Size is the size as a percentage of BERT-base, and Speedup is the inference speedup over BERT-base.}
\label{tab:peft}
\small
\setlength{\tabcolsep}{4pt}
\renewcommand{\arraystretch}{1.15}
\begin{tabular}{ll|ccc|cc}
\hline
\textbf{Model} & \textbf{Variant} & \textbf{POS} & \textbf{NER} & \textbf{EQA}
& \textbf{Size} & \textbf{Speedup} \\
\hline
\multirow{2}{*}{DistilBERT}
 & + LoRA        & 73 & 6  & 26 & 60.30 & \textbf{1.55} \\
 & + LoRA + AlAd & \textbf{96} & \textbf{90} & \textbf{58} & 71.08 & 1.47 \\
\hline
\multirow{2}{*}{ASC}
 & + LoRA        & 82 & 24 & 12 & 48.30 & \textbf{2.04} \\
 & + LoRA + AlAd & \textbf{98} & \textbf{76} & \textbf{69} & 59.08 & 1.89 \\
\hline
\multirow{2}{*}{BERT-M}
 & + LoRA        & 72 & 10 & 3  & 10.25 & \textbf{2.48} \\
 & + LoRA + AlAd & \textbf{92} & \textbf{71} & \textbf{34} & 15.03 & 2.28 \\
\hline
\multirow{2}{*}{BERT-T}
 & + LoRA        & 59 & 2  & 0 & 4.03 & \textbf{2.65} \\
 & + LoRA + AlAd & \textbf{82} & \textbf{51} & \textbf{1} & 7.31 & 2.15 \\
\hline
\end{tabular}
\end{table}

\subsection{AlAd with Parameter Efficient Fine-Tuning}
AlAd does not compete with PEFT methods such as LoRA. It can work in conjunction with LoRA. Table~\ref{tab:peft} reports the results when the
compressed model is kept frozen and adapted with LoRA. LoRA on its own is not effective in this setting. It performs worse than the fine-tuned compressed model for every model and every task. This lower performance of LoRA is expected. LoRA trains a small number of parameters in the attention layers and leaves the rest of the model unchanged. It therefore assumes that the frozen model already produces good representations, and that only a small task-specific correction is needed. A compressed model does not satisfy this assumption. The representation itself has degraded, and a low-rank update to the attention layers cannot recover it. Adding AlAd improves every
model with LoRA on every task. For POS tagging, the accuracy of ASC rises from 82 to 98. For NER, the macro F1 of BERT-T rises from 2 to 51. The gain on EQA is the largest. The exact match of ASC rises from 12 to 69, and that of DistilBERT rises from 26 to 58. AlAd helps here because it modifies the representation rather than the task adaptation. A practitioner who chooses LoRA over full fine-tuning therefore benefits from adding AlAd. The only exception is BERT-T on EQA, where neither configuration is practically usable.

\begin{table}[t]
\centering
\caption{Results when the compressed model is kept frozen, compared against the fine-tuned compressed model. All performance values are a percentage of the performance of BERT-base on the corresponding metric. Size is the size as a percentage of BERT-base, and Speedup is the inference speedup over BERT-base.}
\label{tab:frozen}
\small
\setlength{\tabcolsep}{4pt}
\renewcommand{\arraystretch}{1.15}
\begin{tabular}{ll|ccc|cc}
\hline
\textbf{Model} & \textbf{Variant} & \textbf{POS} & \textbf{NER} & \textbf{EQA}
& \textbf{Size} & \textbf{Speedup} \\
\hline
\multirow{2}{*}{DistilBERT}
 & fine-tuned      & \textbf{100} & 88 & \textbf{89} & 60.00 & \textbf{1.60} \\
 & frozen + AlAd   & 95  & \textbf{94} & 7  & 65.89 & 1.54 \\
\hline
\multirow{2}{*}{ASC}
 & fine-tuned      & 96 & 53 & \textbf{68} & 48.00 & \textbf{2.12} \\
 & frozen + AlAd   & \textbf{98} & \textbf{75} & 34 & 53.89 & 2.01 \\
\hline
\multirow{2}{*}{BERT-M}
 & fine-tuned      & 92 & 27 & \textbf{61} & 10.20 & \textbf{2.50} \\
 & frozen + AlAd   & 92 & \textbf{71} & 3  & 12.68 & 2.41 \\
\hline
\multirow{2}{*}{BERT-T}
 & fine-tuned      & 79 & 8  & \textbf{20} & 4.01 & \textbf{2.80} \\
 & frozen + AlAd   & \textbf{86} & \textbf{57} & 1  & 5.70 & 2.27 \\
\hline
\end{tabular}
\end{table}

\subsection{AlAd and the Difficulty of the Task}

The three tasks need different amounts of intervention. AlAd is more useful for tagging oriented tasks such as POS tagging and NER as compared to the span oriented task of EQA. 

NER is the task on which compression costs the most. Please refer to Table~\ref{tab:main}. In simple fine-tuning configuration, the macro F1 of BERT-M falls to 27 and that of BERT-T falls to 8. It is also the task on which AlAd recovers the most. AlAd in joint fine-tuning mode raises these to 80 and 59. Macro F1 weights every entity type equally. Compression therefore affects the rare entity types the most, and AlAd recovers those types the most.

For NER, a frozen compressed model with AlAd is sufficient.
Please refer to Table~\ref{tab:frozen}. It compares AlAd with frozen compressed model against the baseline of fine-tuned compressed model. AlAd in this configuration still performs better than the fine-tuned compressed model on NER for all four models. It reaches 71 for BERT-M against 27, and 57 for BERT-T against 8. The fine-tuned model is the stronger baseline here. When a model is fine-tuned for the task, its all parameters are updated for the task. In contrast, AlAd with frozen compressed model does not change a single parameter of the compressed model while adapting to the task. A practitioner who cannot retrain the compressed model can therefore still improve it substantially. For
POS tagging the frozen mode is also competitive. It performs better than the fine-tuned model for ASC and BERT-T, it matches it for BERT-M, and it is lower for DistilBERT.

For EQA the frozen compressed model mode of AlAd is not sufficient. It performs worse than the fine-tuned
compressed model for all four models. Please refer to Table~\ref{tab:frozen}. The exact match of BERT-M falls from 61 to 3, and that of DistilBERT falls from 89 to 7. AlAd in the joint fine-tuning mode is necessary for this task (Table~\ref{tab:main}). Span extraction requires the answer boundaries to be located in the
input. Aligning the token embeddings changes the representation, but the model
that produces it is unchanged, and the boundaries are not located correctly. In terms of the performance improvement, EQA is the hardest task for AlAd.
\begin{table}[b]
\centering
\caption{Larger window size of AlAd generally leads to better performance improvement. All performance values are a percentage of the performance of BERT-base on the corresponding metric. Size is the size as a percentage of BERT-base, and Speedup is the inference speedup over BERT-base.}
\label{tab:window}
\small
\setlength{\tabcolsep}{4pt}
\renewcommand{\arraystretch}{1.15}
\begin{tabular}{ll|ccc|cc}
\hline
\textbf{Model} & \textbf{Adapter} & \textbf{POS} & \textbf{NER} & \textbf{EQA}
& \textbf{Sz} & \textbf{Spd} \\
\hline
\multirow{4}{*}{DistilBERT}
 & Linear & 100 & 102 & 91 & 61.48 & \textbf{1.58} \\
 & W-1    & \textbf{101} & 104 & 91 & 62.28 & 1.55 \\
 & W-3    & \textbf{101} & \textbf{106} & 92 & 64.09 & 1.54 \\
 & W-5    & \textbf{101} & \textbf{106} & \textbf{93} & 65.89 & 1.54 \\
\hline
\multirow{4}{*}{ASC}
 & Linear & 99  & 75 & 70 & 49.01 & \textbf{2.10} \\
 & W-1    & 99  & 76 & 70 & 50.28 & 2.03 \\
 & W-3    & \textbf{100} & 76 & 74 & 52.09 & 2.02 \\
 & W-5    & \textbf{100} & \textbf{80} & 74 & 53.89 & 2.01 \\
\hline
\multirow{4}{*}{BERT-M}
 & Linear & 96 & 51 & 59 & 10.45 & \textbf{2.47} \\
 & W-1    & 97 & 69 & 65 & 11.48 & 2.46 \\
 & W-3    & \textbf{98} & 78 & \textbf{68} & 12.08 & 2.43 \\
 & W-5    & \textbf{98} & \textbf{80} & 65 & 12.68 & 2.41 \\
\hline
\multirow{4}{*}{BERT-T}
 & Linear & 86 & 24 & 20 & 4.13 & \textbf{2.60} \\
 & W-1    & 88 & 45 & 15 & 5.09 & 2.29 \\
 & W-3    & 90 & 53 & 32 & 5.40 & 2.28 \\
 & W-5    & \textbf{91} & \textbf{59} & \textbf{35} & 5.70 & 2.27 \\
\hline
\end{tabular}
\end{table}

\subsection{AlAd and the Effect of the Window Size}

Table~\ref{tab:window} reports the effect of the adapter type and the window
size. Performance improves as the window size grows. Between W-1 and W-5, the
NER macro F1 of BERT-T rises from 45 to 59 and its exact match on EQA rises from
15 to 35. The effect is smaller for the less compressed models. For DistilBERT,
the POS accuracy is unchanged at 101 across the three window sizes, and the NER
macro F1 rises by two points. There are two cases where a wider window does not
help. BERT-M reaches 68 on EQA at W-3 and 65 at W-5, and DistilBERT does not
improve on POS tagging beyond W-1.

The comparison with the linear adapter separates the two components of AlAd. The
linear adapter differs from W-1 only in the hidden layer and the nonlinear
activation, and W-1 differs from W-5 only in the window. For DistilBERT and ASC,
the linear adapter and W-1 perform almost identically, and the hidden layer
contributes at most two points. For BERT-M and BERT-T, the hidden layer
contributes much more. The NER macro F1 rises by 18 points for BERT-M and by 21
points for BERT-T. DistilBERT and ASC are both derived from BERT-b, while BERT-M
and BERT-T are pretrained independently. A linear transformation is therefore
sufficient when the two models share a lineage, and it is not sufficient when
they do not. A similar result has been reported for convolutional networks,
where a single affine layer matches the representations of two networks that
share an architecture and differ only in
initialisation~\cite{csiszarik2021similarity}.

The two components also differ in what they cost. Widening the window increases
the number of parameters. For ASC it adds 3.6 points of the size of BERT-b
between W-1 and W-5, and for BERT-T it adds 0.6 points. The effect on latency is
small. The speedup of BERT-T falls from 2.29 to 2.27 across this range, and that
of ASC falls from 2.03 to 2.01. The hidden layer is the more expensive
component. Adding it reduces the speedup of BERT-T from 2.60 to 2.29 and that of
ASC from 2.10 to 2.03. Once the hidden layer is present, a wider window is an
inexpensive way to improve the alignment.


\begin{table}[hbt]
\centering
\caption{AlAd gains performance at the cost of size and speedup. Performance improvement, size increase and speedup loss are relative to the same compressed model without AlAd.}
\label{tab:delta}
\small
\setlength{\tabcolsep}{5pt}
\renewcommand{\arraystretch}{1.15}
\begin{tabular}{l|ccc|cc}
\hline
\multirow{2}{*}{\textbf{Model}} &
\multicolumn{3}{c|}{\textbf{Performance improvement}} &
\multirow{2}{*}{\shortstack{\textbf{Size}\\\textbf{increase}}} &
\multirow{2}{*}{\shortstack{\textbf{Speedup}\\\textbf{loss}}} \\
\cline{2-4}
& \textbf{POS} & \textbf{NER} & \textbf{EQA} & & \\
\hline
DistilBERT & $1$  & $18$ & $4$  & 9.8\%  & 3.8\% \\
ASC        & $4$  & $27$ & $6$  & 12.3\% & 5.2\% \\
BERT-M     & $5$  & $53$ & $4$  & 24.3\% & 3.6\% \\
BERT-T     & $12$ & $51$ & $15$ & 42.1\% & 18.9\% \\
\hline
\end{tabular}
\end{table}
\begin{table}[hbt]
\centering
\caption{A compressed model with AlAd is a competitive model when compared with next larger model without AlAd. Performance and size of each compressed model with AlAd is expressed as a percentage of performance and size of the next
larger model without AlAd. A performance value above 100 means that the smaller
model with AlAd performs better than the larger model. A size below 100 means it
is smaller. A speedup above 1 means smaller model with AlAd is faster.}
\label{tab:ladder}
\small
\setlength{\tabcolsep}{3.5pt}
\renewcommand{\arraystretch}{1.15}
\begin{tabular}{ll|ccc|cc}
\hline
\textbf{With AlAd} & \textbf{Against} & \textbf{POS} & \textbf{NER} &
\textbf{EQA} & \textbf{Size} & \textbf{Speedup} \\
\hline
DistilBERT & BERT-base     & \textbf{101} & \textbf{106} & 93 & 66 & \textbf{1.54} \\
ASC        & DistilBERT & \textbf{100} & 91  & 83 & 90 & \textbf{1.26} \\
BERT-M     & ASC        & \textbf{102} & \textbf{152} & 96 & 26 & \textbf{1.14} \\
BERT-T     & BERT-M     & 99  & \textbf{214} & 58 & 56 & 0.91 \\
\hline
\end{tabular}
\end{table}

\subsection{AlAd as a Favorable Trade-off}
\label{sec:favorabletradeoff}
Table~\ref{tab:delta} reports the change from each compressed model to the same model with AlAd. The performance gain generally grows as compression increases. BERT-T gains 12 points on POS tagging and 51 points on NER, while DistilBERT gains 1 point and 18 points. The cost grows in the same direction. BERT-T grows by 42 percent of its own parameter count and loses 19 percent of its speedup, while
DistilBERT grows by 10 percent and loses 4 percent. The models that gain the most also pay the most.

Table~\ref{tab:ladder} places this cost in context. It compares each compressed model with AlAd against the next larger model without it. For example, BERT-M with AlAd is a quarter of the size of ASC and is faster than it. It performs better on POS tagging and on NER, and it is competitive on EQA. DistilBERT with AlAd is two thirds of the size of BERT-base and is 1.5 times faster. It performs better on POS tagging and on NER. The remaining two pairs are mixed. ASC with AlAd is smaller and faster than DistilBERT and matches it on POS tagging, but it is lower on NER and on EQA. BERT-T with AlAd is half the size of BERT-M and performs much better on NER, but it is lower on EQA and it is also slower.

The two tables together describe the trade-off. Against its own baseline, a compressed model with AlAd is larger and slower. Against the next larger model, it is smaller and usually faster at the same level of performance on the tagging tasks. 


\section{Related Work}

\begin{table*}[hbt]
\centering
\caption{Comparison of paradigms for improving the performance of a compressed
model. AlAd is the only one that improves a compressed model it did not produce,
without requiring a weight-level correspondence to an uncompressed model, and
that is retained at inference.}
\label{tab:paradigms}
\small
\begin{tabular}{l|l|c|l|l}
\hline
\textbf{Paradigm} & \textbf{Compressed model} & \textbf{Weight correspondence}
& \textbf{Optimisation target} & \textbf{Module at inference} \\
\hline
Knowledge Distillation & Produced by the method & No & Teacher outputs or
features & Discarded \\
PEFT & Given & No & Task loss & Retained or merged \\
Post-training repair & Given & Yes & Original layer output & None \\
Representation alignment & Given & No & Task loss of stitched model & Not
deployed \\
\hline
AlAd & Given & No & Reference model embeddings & Retained \\
\hline
\end{tabular}
\end{table*}

We categorise existing approaches for improving the performance of a compressed model into four paradigms. First, Knowledge Distillation (KD) trains a small model under the supervision of a large model. Second, Parameter Efficient Fine-Tuning (PEFT) adapts a frozen model to a downstream task with a small number of trainable parameters. Third, Post-training repair adjusts the weights of a compressed model after
it has been produced. And fourth, Representation alignment learns a transformation between the representation spaces of two models. We discuss each paradigm below. For each paradigm, we state why it does not address the post-hoc setting.

KD recovers the performance of a compressed model by training it under the supervision of a large model~\cite{hinton2015distilling}. The compressed model
is called the student and the large model is called the teacher. Early methods
matched only the output logits of the two models~\cite{hinton2015distilling}.
Feature-based methods also match the intermediate representations, and they work
better at high compression ratios. FitNets introduced a regressor that maps a
hidden layer of the student to a hidden layer of the teacher, which allows the
two models to differ in width~\cite{romero2015fitnets}. TinyBERT and Patient
Knowledge Distillation apply this idea to transformers by matching hidden states
and attention maps~\cite{jiao-etal-2020-tinybert,sun2019patient}.
Relation-based methods instead transfer the relationships between
representations. MiniLM distils the self-attention relations of the last
transformer layer~\cite{wang2020minilm}. Distillation may also be
task-agnostic. MobileBERT distils a general-purpose compact model that is later
fine-tuned for any task~\cite{sun-etal-2020-mobilebert}. When the capacity gap
between the two models is large, an intermediate model is used to bridge
it~\cite{mirzadeh2020improved}. DistilBERT is a widely used model produced by
distillation~\cite{sanh2019distilbert}. These methods differ in the knowledge
that is transferred and in the scheme used to transfer
it~\cite{gou2021knowledge}. They share one assumption. The compressed model is
produced under the teacher's supervision from the beginning. FitNets also trains
its regressor jointly with the student and discards it once training is
complete~\cite{romero2015fitnets}. KD therefore produces a compressed model, but
it does not improve one that already exists. AlAd differs in both respects. It
takes an already trained compressed model as its input, and the adapter is part
of the deployed model. AlAd therefore does not compete with KD. It can be
applied to a model that KD has already produced, and we report such results for
DistilBERT.

PEFT adapts a frozen model to a downstream task by training a small number of
additional parameters. Adapters insert bottleneck layers between transformer
blocks and train only those layers~\cite{houlsby2019parameter}. LoRA injects
low-rank matrices into the attention layers~\cite{hulora}, and AdaLoRA allocates
the rank across layers according to their importance~\cite{zhang2023adaptive}.
Prompt tuning and prefix tuning instead learn continuous vectors that are
prepended to the input~\cite{lester-etal-2021-power,li-liang-2021-prefix}.
Selective methods train a small subset of the existing parameters, such as the
bias terms~\cite{ben-zaken-etal-2022-bitfit}. QLoRA combines PEFT with
quantization and fine-tunes a quantized model with low-rank
adapters~\cite{dettmers2023qlora}. These methods differ in where the trainable
parameters are placed~\cite{lialin2023scaling}. They share one objective. Every
PEFT method optimises a task loss, and none of them uses a reference model. PEFT
therefore adapts a compressed model to a task, but it does not recover what
compression removed. AlAd optimises the distance to the representations of a
larger model instead of a task loss. PEFT also has one advantage over AlAd. The low-rank matrices of LoRA can be merged into the backbone weights, so LoRA adds no parameters at inference~\cite{hulora}. AlAd cannot be merged, because it maps between two different representation spaces. This is a cost that we measure and report. AlAd can be used in conjunction with PEFT methods. AlAd can be added to a compressed model that already uses PEFT, and we report such results for LoRA.

Post-training repair adjusts a compressed model after it has been produced, without retraining it. Calibration methods correct the numerical error introduced by quantization~\cite{kim2021bert}. ZeroQuant and AWQ recalibrate activations or rescale weights to recover
accuracy~\cite{10.5555/3600270.3602240,MLSYS2024_42a452cb}. Reconstruction methods go further and adjust the compressed weights themselves. AdaRound learns rounding decisions that minimise the difference between the output of a quantized layer and the output of the original layer~\cite{nagel2020adaround}. BRECQ performs this reconstruction block by block~\cite{li2021brecq}, and GPTQ
applies it to large language models~\cite{frantar2023gptq}. SparseGPT solves for weights that reconstruct the output of the dense layer it replaces after pruning~\cite{frantar2023sparsegpt}. These methods share a requirement. Each one repairs a compressed model layer by layer towards the model it was derived from, so it needs a weight-level correspondence between the two models. This correspondence does not always exist. BERT-Mini and BERT-Tiny are pretrained
independently and are not derived from BERT-base, so there is no original layer to reconstruct towards. AlAd does not require this correspondence. It uses only the output embeddings of the compressed model and of the reference model, and it treats both as black boxes.

Representation alignment methods learn a transformation between the
representation spaces of two models. Lenc and Vedaldi introduced a trainable stitching layer that connects the lower part of one network to the upper part of another~\cite{lenc2015understanding}. If the stitched network performs well, the two representation spaces are taken to be compatible. Bansal et al. use stitching to compare the representations learned by different networks~\cite{bansal2021revisiting}. Csiszárik et al. show that for
convolutional networks that share an architecture but differ in initialisation, a single affine layer is often sufficient to match their
representations~\cite{csiszarik2021similarity}. Moschella et al. learn a
transformation that allows the latent space of one model to be used by
another~\cite{moschella2023relative}. Hernandez et al. report that a stitching layer does not always reproduce the representations that the receiving layer expects, and that it may instead find different representations that work as well~\cite{hernandez2023stitching}. These methods learn the same kind of transformation that AlAd learns. They share one purpose. The transformation is an instrument for comparing two models, and the stitched network is a diagnostic rather than a model meant for deployment. Csiszárik et al. measure the accuracy
of the stitched network and then discard it~\cite{csiszarik2021similarity}. We instead keep the stitched model and deploy it, and we report what that costs in parameters and in latency.

Table~\ref{tab:paradigms} summarises these four paradigms. No single property separates AlAd from all of them. AlAd shares its alignment objective with KD, its modular structure with PEFT, and its post-hoc application with post-training repair. What the post-hoc setting requires is the combination. AlAd improves a compressed model that it did not produce, it does not need a weight-level correspondence with an uncompressed model, and the adapter is retained at inference. To the best of our knowledge, existing approaches do not simultaneously address the three requirements of the post-hoc setting considered in our work.

\section{Conclusion and Future Work}

We asked how much of the performance of a reference model can be recovered after
compression, and at what cost. For tagging tasks, most of it can be recovered.
ASC reaches the POS accuracy of BERT-base at half its size, and DistilBERT exceeds
it. For span extraction, a gap remains at every compression ratio. The cost is
paid in parameters and in latency, and it grows as compression grows. The models
that gain the most from alignment are also the ones for which the adapter is
proportionally largest.

Our results suggest a decision procedure for a practitioner who is given a compressed model. If the model can be trained, joint fine-tuning with AlAd is the best option, and it applies whatever produced the compressed model. If the model can only be run in the forward direction, alignment alone is sufficient for tagging tasks and is not sufficient for span extraction. If a PEFT method is already in use, AlAd should be added to it rather than replacing it. Above all, the adapter is worth its overhead when the alternative is a larger model. A compressed model with AlAd is competitive with the next larger model on the tagging tasks, at a smaller size.


Our work can be improved in many ways. Our experiments do not yet separate the contribution of the sliding window from the contribution of the parameters it adds, and a parameter-matched comparison would settle this. We also do not isolate the contribution of the alignment phases. Finally, our experiments are limited to encoder-only models. Our three tasks assign a label or a span to positions in the input, and encoders are the standard architecture for them. Whether post-hoc alignment behaves in the same way for decoder-only models is an open question.

\bibliographystyle{ACM-Reference-Format}
\bibliography{sample-base}


\end{document}

%% file: figures/alad-figure.tex
%

\begin{figure*}[t]
\centering
\begin{subfigure}[b]{0.26\textwidth}
\centering
\begin{tikzpicture}
  \foreach \i in {0,...,6} \node[emb] (r\i) at (\i*0.52,1.55) {};
  \node[emb,fill=black!12] (rp) at (-0.52,1.55) {0};
  \node[tny,above=0.5mm of r1] {$W_{i-2}$};
  \node[tny,above=0.5mm of r3] {$W_{i}$};
  \node[tny,above=0.5mm of r5] {$W_{i+2}$};
  \node[lbl,left=1mm of rp] {$R_C$};
  \begin{scope}[on background layer]
    \node[fill=blue!13,rounded corners=2pt,fit=(r1)(r5),inner sep=1.5pt] (win) {};
  \end{scope}
  \node[trn,minimum width=1.4cm] (alad) at (1.56,0.6) {AlAd};
  \foreach \i in {1,...,5} \draw[ar,black!50,line width=0.5pt] (r\i) -- (alad);
  \node[emb,fill=blue!30] (out) at (1.56,-0.15) {};
  \draw[ar] (alad) -- (out);
  \node[lbl,right=1.5mm of out] {$R'_{Ci}$};
  
\end{tikzpicture}
\caption{Sliding window ($n=5$)}\label{fig:a}
\end{subfigure}
\hfill
\begin{subfigure}[b]{0.40\textwidth}
\centering
\begin{tikzpicture}
  \foreach \p/\y in {1/2.0,2/1.15,3/0.3} \node[lbl,anchor=east] at (-0.1,\y) {Phase \p};
  \node[froz,minimum width=1.0cm] (c1) at (0.72,2.0) {$M_C$};
  \node[trn,minimum width=1.0cm,right=2mm of c1] (a1) {AlAd};
  \node[lbl,right=2mm of a1,anchor=west,align=left] {MSE to $R_L$, general corpus};
  \draw[ar] (c1)--(a1);
  \node[froz,minimum width=1.0cm] (c2) at (0.72,1.15) {$M_C$};
  \node[trn,minimum width=1.0cm,right=2mm of c2] (a2) {AlAd};
  \node[lbl,right=2mm of a2,anchor=west,align=left] {MSE to $R_L$, task data};
  \draw[ar] (c2)--(a2);
  \node[blk,fill=black!8,draw=blue!65,minimum width=1.0cm] (c3) at (0.72,0.3) {$M_C$};
  \node[trn,minimum width=1.0cm,right=2mm of c3] (a3) {AlAd};
  \node[trn,minimum width=0.85cm,right=2mm of a3] (h3) {Head};
  \node[lbl,right=2mm of h3,anchor=west] {task loss};
  \draw[ar] (c3)--(a3); \draw[ar] (a3)--(h3);
  \node[tny,anchor=west] at (-0.75,-0.35) {%
    \tikz\node[froz,minimum width=3mm,minimum height=3mm,inner sep=0pt]{};\ frozen \quad
    \tikz\node[trn,minimum width=3mm,minimum height=3mm,inner sep=0pt]{};\ trained \quad
    \tikz\node[blk,fill=black!8,draw=blue!65,minimum width=3mm,minimum height=3mm,inner sep=0pt]{};\ frozen \emph{or} trained};
\end{tikzpicture}
\caption{Three training phases}\label{fig:b}
\end{subfigure}
\hfill
\begin{subfigure}[b]{0.31\textwidth}
\centering
\begin{tikzpicture}
  \node[blk,minimum width=0.95cm] (l1) at (0,2.0) {$M_L$};
  \node[blk,minimum width=0.85cm,right=2mm of l1] (h1) {Head};
  \draw[ar] (l1)--(h1);
  \node[blk,minimum width=0.95cm] (c2) at (0,1.0) {$M_C$};
  \node[blk,minimum width=0.85cm,right=2mm of c2] (h2) {Head};
  \draw[ar] (c2)--(h2);
  \node[blk,minimum width=0.95cm] (c3) at (0,0.0) {$M_C$};
  \node[trn,minimum width=0.95cm,right=2mm of c3] (a3) {AlAd};
  \node[blk,minimum width=0.85cm,right=2mm of a3] (h3) {Head};
  \draw[ar] (c3)--(a3); \draw[ar] (a3)--(h3);
  \begin{scope}[on background layer]
    \node[draw=blue!60,thick,rounded corners=2.5pt,fit=(c3)(a3)(h3),inner sep=2.5pt] (dep) {};
  \end{scope}
\end{tikzpicture}
\caption{Three deployment options}\label{fig:c}
\end{subfigure}

\caption{Details of the Alignment Adapter (AlAd). (a) AlAd reads a window of
$n$ neighbouring embeddings of the compressed model $M_C$ and produces a single
embedding $R'_{Ci}$ in the representation space of the reference model $M_L$;
the sequence is zero-padded at the boundaries. (b) AlAd is trained in three
phases. The first two align $R'_{C}$ with $R_{L}$ while $M_C$ remains frozen.
The third trains AlAd and a task-specific inference module with a task loss, and
$M_C$ may be kept frozen or fine-tuned jointly. (c) Three deployment options.
$M_L$ is the largest and the slowest, and it performs the best. $M_C$ is the smallest
and the fastest, and its performance is the lowest. $M_C$ with AlAd lies between them and
offers a favorable trade-off between efficiency and performance.}
\label{fig:alad-overview}
\end{figure*}